\documentclass[paper]{ieice}
\usepackage[pdftex]{graphicx,xcolor}
\usepackage[fleqn]{amsmath}
\usepackage{newtxtext}
\usepackage[varg]{newtxmath}

\usepackage{url}
\usepackage{pgfplots}
\pgfplotsset{width=10cm,compat=1.9}
\usepackage{CJK}
\usepackage{multirow} 
\usepackage{color}

\newcommand{\etal}{\textit{et al. }}

\field{}
\title{Document-level Neural Machine Translation with Associated Memory Network}
\authorlist{%
 \authorentry{Shu Jiang}{}{label1,label2}\MembershipNumber{}
 \authorentry{Rui Wang}{}{label1,label2}\MembershipNumber{}
 \authorentry{Zuchao Li}{}{label1,label2}\MembershipNumber{}
 \authorentry{Masao Utiyama}{}{label3}\MembershipNumber{}
 \authorentry{Kehai Chen}{}{label3}\MembershipNumber{}
 \authorentry{Eiichiro Sumita}{}{label3}\MembershipNumber{}
 \authorentry[zhaohai@cs.sjtu.edu.cn (Corresponding author.)]{Hai Zhao}{}{label1,label2}\MembershipNumber{}
 \authorentry[bllu@sjtu.edu.cn (Corresponding author.)]{Bao-liang Lu}{}{label1,label2}\MembershipNumber{}
}
\affiliate[label1]{The authors are with the Department of Computer Science and Engineering, Shanghai Jiao Tong University, Shanghai, 200240, China.}
\affiliate[label2]{The authors are with the Key Laboratory of Shanghai Education Commission for Intelligent Interaction and Cognitive Engineering, Shanghai Jiao Tong University, Shanghai, 200240, China.}
\affiliate[label3]{The authors are with National Institute of Information and Communications Technology, Kyoto-shi, 619–0289 Japan.}

\begin{document}
\begin{CJK*}{UTF8}{gbsn}

\maketitle
\begin{summary}
Standard neural machine translation (NMT) is on the assumption that the document-level context is independent. Most existing document-level NMT approaches are satisfied with a smattering sense of global document-level information, while this work focuses on exploiting detailed document-level context in terms of a memory network. The capacity of the memory network that detecting the most relevant part of the current sentence from memory renders a natural solution to model the rich document-level context. In this work, the proposed document-aware memory network is implemented to enhance the Transformer NMT baseline. Experiments on several tasks show that the proposed method significantly improves the NMT performance over strong Transformer baselines and other related studies.
\end{summary}

\section{Introduction}

Neural Machine Translation (NMT) \cite{Kalchbrenner2013recurrent,Sutskever2014Advances,Cho2014Learning,bahdanau2015attention,vaswani2017attention} established on the encoder-decoder framework, where the encoder takes a source sentence as input and encodes it into a fixed-length embedding vector, and the decoder generates the translation sentence according to the encoder embedding, has achieved advanced translation performance in recent years.
So far, most models take a standard assumption to translate every sentence independently, ignoring the document-level contextual clues during translation.

% 1 为什么Document在NMT中重要，说明你的课题有意义。 
However, document-level information can improve the translation performance from multiple aspects: consistency, disambiguation, and coherence \cite{kuang2018cache}. If translating every sentence is independent of the document-level context, it will be challenging to keep every sentence translation across the entire text consistent with each other. Moreover, the document-level context can also assist the model to disambiguate words with multiple senses, and the global context is of great benefit to translation in a coherent way.

% 2 已有方法如何利用Document，有何优缺点。这里可以简要介绍已有方法，不用公式。
There have been few recent attempts to introduce the document-level information into the existing standard NMT models.
Various existing methods \cite{Jean2017context,tiedemann2017context,wang2017exploiting,voita2018context,kuang2018gate} focus on modeling the context from the surrounding text in addition to the source sentence.
% \cite{wang2017exploiting} use a hierarchical Recurrent Neural Network (RNN) to import the information of previous sentences.
For the more high-level context, Miculicich \etal \cite{miculicich2018han} propose a multi-head hierarchical attention machine translation model to capture the word-level and sentence-level information.
The cache-based model raised by Kuang \etal \cite{kuang2018cache} uses the dynamic cache and topic cache to capture the inter-sentence connection.
Tan \etal \cite{tan-etal-2019-hierarchical} propose a hierarchical model of global document context to improve document-level translation.
In addition, many studies \cite{wang2017exploiting,kuang2018gate,voita2018context} all add the contextual information to the NMT model by applying the gating mechanism \cite{tu2017context} to dynamically control the auxiliary global context information at each decoding step. 

However, most of the existing document-level NMT methods focus on briefly introducing the global document-level information but fail to consider selecting the most related part inside the document context.
% most of the existing document-level NMT methods have to inconveniently prepare the contextual input or model the global context in advance. 
% Another potential limitation is that they are limited by the context sentence

% 3 针对已有方法的缺点，本文提出一种XXX方法，特点和优点是什么
Inspired by the observation that human and document-level machine translation models always refer to the source sentence's context during the translation, like query in their memory, we propose to utilize the document-level sentences associated with the source sentences to help predict the target sentence. To reach such a goal, we adopt a Memory Network component \cite{Weston2015Memory,sukhbaatar2015end,guan-etal-2019-semantic} which provides a natural solution for the requirement of modeling document-level context in document-level NMT. 
In fact, Maruf and Haffari \cite{maruf2018document} have already presented a document-level NMT model which projects the document contexts into the tiny dense hidden state space for RNN model using memory networks and updates word by word, and their model is effective in exploiting both source and target document context.
% In other words, it predicts a target word not only on the previously generated words and the current source sentence (as in the vanilla NMT model) but also on all the other source sentences of the document and their translations.
% However, the cache in RNN does not decay exponentially and has roughly the same average activation across the entire memory\cite{sukhbaatar2015end}.%, which differs with ours on Transformer NMT memorizing complete contextual sentences.

Different from any previous work, this paper presents a Transformer NMT model with document-level Memory Network enhancement \cite{Weston2015Memory,sukhbaatar2015end} which concludes contextual clues into the encoder of the source sentence by the Memory Network. Not like Maruf and Haffari \cite{maruf2018document} which memorizes the whole document information into a tiny dense hidden state, the memory in our work calculates the associated document-level contextualized information in the memory with the current source sentence using the attention mechanism.
In this way, our proposed model is able to focus on the most relevant part of the concerned translation from the memory, which precisely encodes the concerned document-level context.

The empirical results indicate that our proposed method significantly improves the BLEU score compared with a strong Transformer baseline and performs better than other related models for document-level machine translation on multiple language pairs with multiple domains.

% 主图 
\begin{figure*}[tb]
\begin{center}
\includegraphics[width=1.0\textwidth]{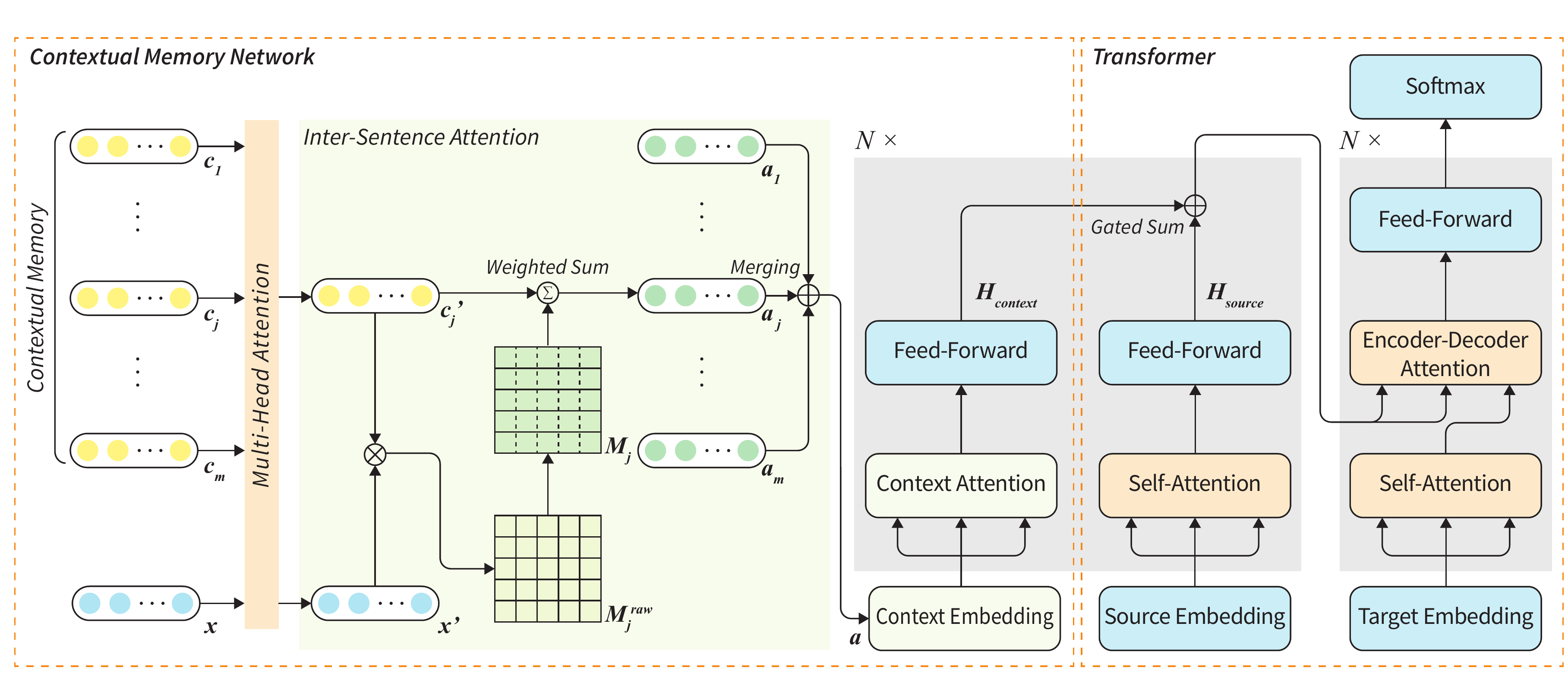}
\end{center}
\caption{The framework of our model.}
\label{fig:framework}
\end{figure*}

\section{Background}
\subsection{Neural Machine Translation}
Given a source sentence \textcolor{black}{with $S$ tokens} $\mathbf{x} = \{x_{1}, ..., x_{i}, ... , x_{S}\}$ in the document to be translated and a target sentence \textcolor{black}{with $T$ tokens} $\mathbf{y}= \{y_{1}, ..., y_{i}, ... , y_{T}\}$, NMT model computes the probability of translation from the source sentence to the target sentence word by word:
\begin{equation}
    P(\mathbf{y}|\mathbf{x}) = \prod^{T}_{i=1}P(y_i|y_{1:i-1},\mathbf{x}),
\end{equation}
where $y_{1:i-1}$ is a substring containing words $y_{1}, ..., y_{i-1}$.
Generally, with an RNN, the probability of generating the $i$-th word $y_{i}$ is modeled as: 
\begin{equation}
    P(y_i|y_{1:i-1},\mathbf{x}) = \mbox{softmax}(g(y_{i-1},\mathbf{s}_{i-1},\mathbf{c}_i)),
\end{equation}
where $g(\cdot)$ is a nonlinear function that outputs the probability of previously generated word $y_i$, and $\mathbf{c}_i$ is the $i$-th source representation.
Then $i$-th decoding hidden state $\mathbf{s}_i$ is computed as 
\begin{equation}
    \mathbf{s}_i = f(\mathbf{s}_{i-1}, y_{i-1}, \mathbf{c}_i).
\end{equation}

For NMT models with an encoder-decoder framework, the encoder maps an input sequence of symbol representations $\mathbf{x}$ to a sequence of continuous representations $\mathbf{z}= \{z_{1}, ..., z_{i}, ... , z_{S}\}$. Then, the decoder generates the corresponding target sequence of symbols $\mathbf{y}$ one element at a time.

\subsection{Transformer Architecture}
Only based on the attention mechanism, a network architecture called Transformer \cite{vaswani2017attention} for NMT uses stacked self-attention and point-wise, fully connected layers for both encoder and decoder. 

% As illustrated in Figure \ref{fig:transformer} (a), the 
A stack of $N$ (usually equals to 6) identical layers constitutes the encoder, and each layer has two sub-layers: (1) multi-head self-attention mechanism, and (2) a simple, position-wise fully connected feed-forward network. 

Multi-head attention 
% demonstrated in the Figure \ref{fig:transformer} (b) 
in the Transformer allows the model to process information jointly from different representation spaces at different positions.
It linearly projects the queries $Q$, keys $K$, and values $V$ $h$ times with different, learned linear projections to $d_k$, $d_k$, and $d_v$ dimensions respectively, and then the attention function is performed in parallel, generating $d_v$-dimensional output values, and yielding the final results by concatenating and once again projecting them. The core of multi-head attention is Scaled Dot-Product Attention and calculated as:
\begin{equation}
    \mbox{Attention}(Q,K,V) = \mbox{softmax}(\frac{QK^T}{\sqrt{d_k}})V.
\end{equation}
% The multi-head attention mechanism is a crucial part of the Transformer architecture, and it is also applied in our proposed method.

The second sub-layer is a feed-forward network containing two linear transformations with a ReLU activation in between.

Similar to the encoder, the decoder is also composed of a stack of $N$ identical layers, but it inserts a third sub-layer, which performs multi-head attention over the output of the encoder stack. The Transformer also employs residual connections around each of the sub-layers, followed by layer normalization. Thus, the Transformer is more parallelizable and faster for translating than earlier RNN methods. 

\subsection{Memory Network}
% mn是lazy learning ，传统模型下也有类似研究。
Memory networks \cite{Weston2015Memory} utilize the external memories as inference components based on long-range dependencies, which can be categorized into a sort of lazy machine learning \cite{aha2013lazy}. Using the similar memorizing mechanism, memory-based learning methods have been also applied in multiple traditional models \cite{daelemans1999introduction,fix1951discriminatory,skousen1989analogical,skousen2013analogy,lebowitz1983memory,nivre2004memory}.
A memory network \cite{Weston2015Memory} is a set of vectors ${\color{black}\mathcal{M}}=\{\mathbf{m}_1,...,\mathbf{m}_K\}$ and the memory cell $\mathbf{m}_k$ is potentially relevant to a discrete object (for example, a word) $x_k$. 
The memory is equipped with a \emph{read} and optionally a \emph{write} operation.
Given a query vector $\mathbf{q}$, the output vector produced by reading from the memory is $\sum^{K}_{i=1} p_i \mathbf{m}_i$,
where $p_i =\mbox{softmax}(\mathbf{q}^T \cdot {\color{black}\mathcal{M}})$ scores the match between the query vector $\mathbf{q}$ and the $i$-th memory cell $\mathbf{m}_i$.

\section{Model}
% \section{Inter-Sentence Attention: How to Introduce Document Information}
%这里说如何引进文档信息

\subsection{Framework}
Our NMT model consists of two components: \textit{Contextual Associated Memory Network} and a Transformer model.
For the \textit{Contextual Associated Memory Network}, the core part is a neural controller, which acts as a ``processor'' to read memory from the contextual storage ``RAM'' according to the input before sending it to other components. The controller calculates the correlation between the input and memory data, i.e., ``memory addressing''.

\subsection{Encoders}
Our model requires two encoders: the source encoder for translation from input sentence representation and the context encoder for the \textit{Contextual Associated Memory Network} from context sentence representation. 
The source encoder is composed of a stack of $N$ layers, the same as the source encoder in the original Transformer \cite{vaswani2017attention}.
The proposed \textit{Contextual Associated Memory Network} consists of four parts: context selection, inter-sentence attention, embedding merging, and context gating.

\subsection{Contextual Associated Memory Network}
For each source sentence $\mathbf{x}$ at each training step, we assume the $m$ context sentences $\{\mathbf{c}_j\}^{m}_{j=1}$ related with the current sentence $\mathbf{x}$ as the \textit{contextual memory} with the memory size $m$.

% 句子选择
\subsubsection{Context Selection}
For the sake of fairness, we can treat all sentences in the document as our memory. However, it is impossible to attend all the sentences in the training dataset because of the extremely high computing and memorizing cost. 
We aim to utilize the context sentences and their representations to help our model predict the target sentences.
There are three common ways to select the context sentences: \textit{previous} sentences of the current sentence, \textit{next} sentences of the current sentence, and \textit{context} sentences randomly selected from the training corpus \cite{voita2018context}. 

\subsubsection{Inter-Sentence Attention}
This part aims to attain the inter-sentence attention matrix, which can also be regarded as the core part of the \textit{Contextual Associated Memory Network}.
The input sentence ${\color{black} \mathbf{x}}$ and the context sentences $\{\mathbf{c}_j\}^{m}_{j=1}$ in the \textit{contextual memory} first go through a multi-head attention layer to encode the word representation:
\begin{equation}
    \mathbf{x'} = \mbox{MultiHead}(\mathbf{x},\mathbf{x},\mathbf{x}) ,
\end{equation}
and
\begin{equation}
    \mathbf{c}_j' = \mbox{MultiHead}(\mathbf{c}_j,\mathbf{c}_j,\mathbf{c}_j) \ \  j\in\{1,2,...,m\} ,
\end{equation}

The lists of new word representations are denoted as follows:
\begin{equation}
\mathbf{x'} = \{x'_{1}, ..., x'_{i}, ... , x'_{S}\},
\end{equation}
and 
\begin{equation}
\mathbf{c}_j' = \{c'_{1,j}, ..., c'_{k,j}, ... , c'_{K_j,j}\}  \ \  j\in\{1,2,...,m\}.
\end{equation}
Each word representation is as a vector $x \in \mathbb{R}^d $, 
where $d$ is the size of hidden state in $\mbox{MultiHead}$ function.

Then, for each context sentence representation $\mathbf{c_j'}$, we apply the multi-head attention by treating the input sentence representation $\mathbf{x'}$ as the query sequence, on them and get the attention matrix ${\color{black}\mathcal{M}}_j^{raw}$:
\begin{equation}
    {\color{black}\mathcal{M}}^{raw}_j = \mathbf{x}' \otimes \mathbf{c}_j'^T  \ \  j\in\{1,2,...,m\}.
\end{equation}

Every element ${\color{black}\mathcal{M}}_{raw}(i,k)=x'_i \cdot c'^{T}_{k,j}$ can be regarded as an indicator of similarity between the $i$-th word in input sentence representation $\mathbf{x}'$ and the $k$-th word in memory sentence representation $\mathbf{c}_j'$.

Finally, we perform a softmax operation on every column in ${\color{black}\mathcal{M}}_j^{raw}$ to normalize the value so that it can be considered as the probability from input sentence representation $\mathbf{x}'$ to memory sentence representation $\mathbf{c}_j'$:
\begin{equation}
    \alpha_{i,j} = \mbox{softmax}([{\color{black}\mathcal{M}}_j^{raw}(i,1),...,[{\color{black}\mathcal{M}}_j^{raw}(i,K_j)]),
\end{equation}
and
\begin{equation}
    {\color{black}\mathbf{M}}_j=[\alpha_{1,j},...,\alpha_{i,j},...,\alpha_{S,j}].
\end{equation}

We treat the probability vector $\alpha_{i,j}$ as a set of weights to sum all the representations in $\mathbf{c}_j'$ and get the memory-sentence-specified argument embedding $\mathbf{a}_{j}$:
\begin{equation}
    \mathbf{a}_{j} = [a_{1,j},...,a_{i,j},...,a_{\color{black}K_j,j}],
\end{equation}
where
% \begin{equation}
%     a_{i,j} = \alpha_{i,j} \cdot \mathbf{c}_j'^{T} 
%     = \sum^{K_j}_{k=1} \alpha_{i,j} c'^{\color{black}T}_{k,j}.
% \end{equation}
\begin{equation}
    {\color{black}a_{i,j} = \sum^{K_j}_{k=1} \alpha_{i,j} c'_{k,j}}.
\end{equation}

\subsubsection{Embedding Merging}
\label{sec:merging}
To utilize the contextual embeddings $\mathbf{a}_{j}$ of the context sentences during training, embedding merging needs to be done.

Because the context sentences are different, the overall contributions of these word representations should be different. 
We let the model itself learn how to make use of these contextual word representations. 
Following the attention combination mechanism \cite{libovicky2017attention,guan-etal-2019-semantic}, we consider four ways to merge the label information.

\paragraph{Concatenation} 
All the contextual argument embedding are concatenated as the final attention embeddings.
\begin{equation}
\mathbf{a} = [\mathbf{a}_{1},...,\mathbf{a}_{j},...,\mathbf{a}_{m}].
\end{equation}

\paragraph{Average}
The average value of all the contextual argument embeddings is used as the final attention embedding.
\begin{equation}
\mathbf{a} = \frac{1}{m}\sum^{m}_{j=1}\mathbf{a}_{j}.
\end{equation}

\paragraph{Weighted Average}
% we use a weighted average strategy to combine these attention representations from different memory sources.
The weighted average of all the contextual argument embedding is used as the final attention embedding.
We calculate the mean value of every raw similarity matrix ${\color{black}\mathcal{M}}_j^{raw}$ to indicate the similarity between input sentence $\mathbf{x}$ and context sentence $\mathbf{c}_j$, and we use the softmax function to normalize them to get a probability vector $\beta$ indicating the similarity of input sentence $\mathbf{x}$ towards all the context sentences $\{\mathbf{c}_j\}_{j=1}^m$:
% \begin{equation}
\begin{align}
\beta & = \mbox{softmax}([g({\color{black}\mathcal{M}}_1^{raw}),...,g({\color{black}\mathcal{M}}_m^{raw})])  \notag \\
      & =[\beta_1,...,\beta_j,...\beta_m],
\end{align}
% \end{equation}
where $g(\cdot)$ represents the mean function. 

Then, we use the probability vector $\beta$ as weight to sum all the contextual attention embedding $a_{i,j}$ for the final contextual attention embedding $\mathbf{a}$ of the $i$-th word $x_i$ in input sentence $\mathbf{x}$:
\begin{equation}
\mathbf{a} = \sum_{j=1}^{m}\beta_j \mathbf{a}_{j}.
\end{equation}

\paragraph{Flat} 
This method does not use $\mathbf{a}_{j}$.
First, we concatenate all the raw similarity matrix ${\color{black}\mathcal{M}}_j^{raw}$ along the row.
\begin{equation}
{\color{black}\mathcal{M}}^{raw} = [{\color{black}\mathcal{M}}_1^{raw},...,{\color{black}\mathcal{M}}_j^{raw},...,{\color{black}\mathcal{M}}_m^{raw}]
\end{equation}
Then, we perform softmax operation on every row in $M^{raw}$ to normalize the value so that it can be considered as probability from input sentence $\mathbf{x}$ to all context sentences $\mathbf{c}_j$ .
\begin{equation}
    \gamma = f([{\color{black}\mathcal{M}}_{1}^{raw},...,{\color{black}\mathcal{M}}_{k}^{raw},...,{\color{black}\mathcal{M}}_{K_{all}}^{raw}]),
\end{equation}
where $f(\cdot)$ stands for softmax operation and $K_{all}$ is the total length of all context sentences, i.e.
\begin{equation}
    K_{all} = \sum^{m}_{j=1} K_j,
\end{equation}
and $K_j$ is the length of context sentences $\mathbf{c}_j$.

We also concatenate the contextual information $\mathbf{c} =  [\mathbf{c}_1,...,\mathbf{c}_j,...,\mathbf{c}_m]$
and use $\gamma$ as weight to sum the concatenated contextual argument embedding as final contextual attention embedding.
\begin{equation}
    \mathbf{a} = \gamma \cdot \mathbf{c}^T.
\end{equation}

\paragraph{Contextual RNN}
We first pad and concatenate the contextual embeddings $\mathbf{a}_{j}$ of the context sentences by columns.
\begin{equation}
    \mathbf{a'} =  [\mathbf{a}_1;...;\mathbf{a}_j;...;\mathbf{a}_m].
\end{equation}

Inspired by the Document RNN method \cite{wang2017exploiting} to summarize the cross-sentence context information, we use RNN by column in the contextual argument embedding to generate the contextual attention embedding, and the hidden state at each time step can represent the relation from the first word embedding to the current word embedding. 

\begin{figure}[ht]
    \centering
    \includegraphics[width=0.48\textwidth]{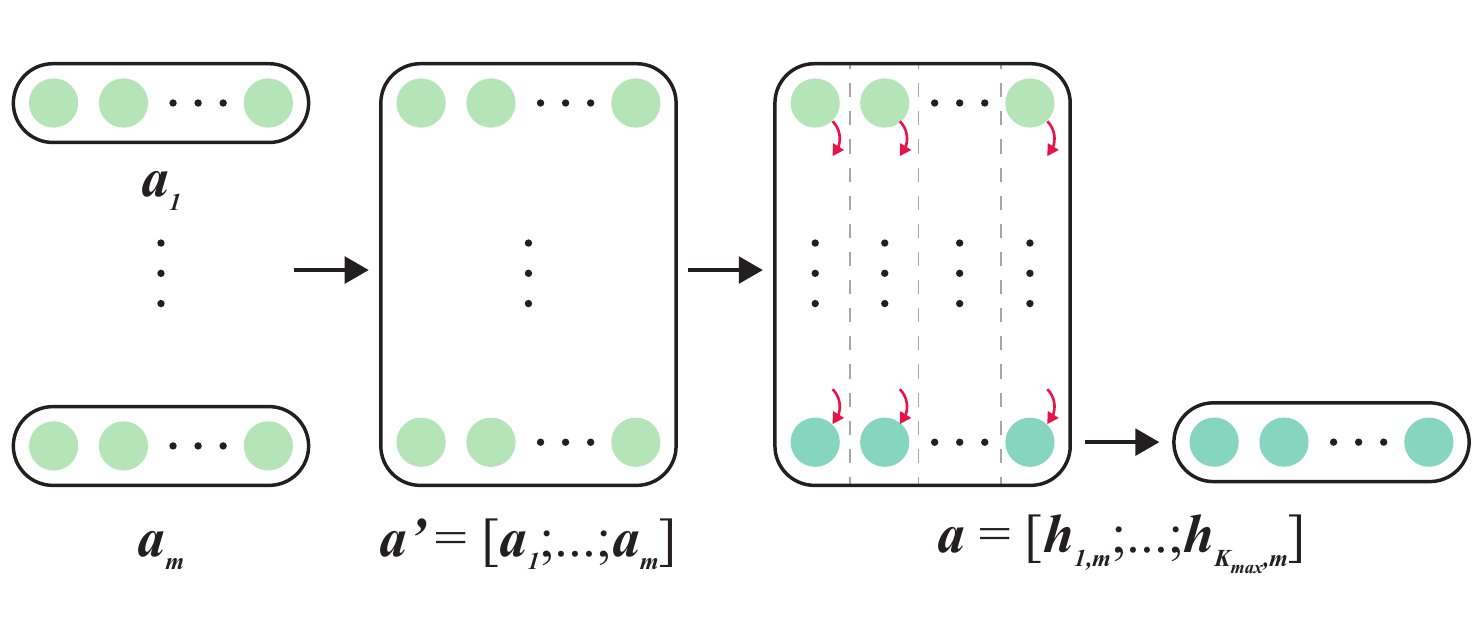}
    \caption{Contextual RNN.}
    \label{fig:rnn}
\end{figure}

As shown in the Figure \ref{fig:rnn}, the RNN output of the $k$-th word embedding $a_{j,k}$ in the contextual argument embedding $a_j$ is 
\begin{equation}
    h_{j,k} = f(h_{j-1,k}, a_{j,k})
\end{equation}
where $f(\cdot)$ is an activation function, and $h_{j,k}$ is the hidden state at time $j$ of the $k$-th word embedding in the contextual argument embedding $a_j$.  

Then we use the hidden state $h_{m,k}$ at the last time $m$, and the final contextual attention embedding $\mathbf{a}$ is concatenated by $h_{m,k}$. 
\begin{equation}
    \mathbf{a} = [h_{m,1},...,h_{m,k},...,h_{m,K_{max}}]
\end{equation}
where $K_{max}$ is the max length of context sentences $\mathbf{c}_j$, i.e. \begin{equation}
    K_{max} =\mathop{\mbox{max}_{j\in\{1,2,...,m\}}}\{K_j\},
\end{equation}
and $K_j$ is the length of context sentences $\mathbf{c}_j$.

\subsubsection{Context Gating}
Since the acquisition of contextual attention embedding $\mathbf{a}$, we operate the MultiHead attention and feed-forward on the contextual attention embedding $a$ and source embedding $\mathbf{x}$ simultaneously like the original Transformer encoding steps, then we annotate the source attention embedding and contextual attention embedding after the above operations as $H_{source}$ and $H_{context}$. To control and analyze the flow of information from the extended context to the translation model, we use a context gate \cite{tu2017context} to integrate the source and context attentions and control the flow from the source side and the context side.
The gate ${g}$ is calculated by 
\begin{equation}
     g = \sigma(W_g[H_{source}, {H}_{context}] + b_g).
\end{equation}
Their gated sum ${H}$ is
\begin{equation}
     H = g \otimes H_{source} + (1 - g) \otimes {H}_{context},
 \end{equation}
where $\sigma$ is the logistic sigmoid function, $\otimes$ is the point-wise multiplication and $W_g$ is trained by the model. As illustrated in Figure \ref{fig:framework}, the output of the gate $H$ is integrated into the encoder-decoder attention part at decoding step.

% 我们做了什么实验
\section{Experimental Setup}
\subsection{Data}
\textcolor{black}{The proposed document-level NMT model will be evaluated on multiple language pairs, i.e., Chinese-to-English (Zh-En), Spanish-to-English (Es-En), French-to-English (Fr-En), and English-to-French (Ja-En) on three domains: talks, subtitles, and news. 
Table \ref{tab:stat} lists the statistics of all the concerned datasets.}

\paragraph{TED Talks}
The Zh-En TED talk documents are the parts of the IWSLT2015 Evaluation Campaign Machine Translation task\footnote{\url{https://wit3.fbk.eu}}. We use \emph{dev2010} as the development set and combine the \emph{tst2010-2013} as the test set.
The Es-En corpus is a subset of the IWSLT2014. We use \emph{dev2010} for development set and \emph{test2010-2012} as the test set.
\textcolor{black}{
The Fr-En corpus is also a subset of the IWSLT2012, where \emph{dev2010} is for development, and \emph{test2010} is the test set.
The Ja-En corpus is from IWSLT2017 and WMT, \emph{dev2010} is for development and \emph{test2010-2015} is for test.}

\paragraph{Subtitles}
The Es-En corpus is a subset of OpenSubtitles2018\footnote{\url{ http://www.opensubtitles.org/}} \cite{lison2016opensub}\footnote{\url{http://opus.nlpl.eu/OpenSubtitles2018.php}}. We randomly select 1,000 continuous sentences for each development set and test set.

\paragraph{News}
The Es-En News-Commentaries11 corpus\footnote{\url{https://opus.nlpl.eu/News-Commentary-v11.php}} has document-level delimitation. We evaluate on the WMT sets \cite{bojar2013WMT}: \emph{newstest2008} for development, and \emph{newstest2009-2013} for testing.

% 数据表格
\renewcommand\arraystretch{1.1}
\begin{table*}[ht]
    \centering
    \caption{Data statistics of sentences.}
    \begin{tabular}{lcccccc}
    \hline
    \multirow{2}{*}{\textbf{Dataset}}  & \multicolumn{4}{c}{\textbf{TED Talks}} & \textbf{Subtitles} & \textbf{News} \\
    \textbf{} & Zh-En    & Es-En    & \textcolor{black}{Fr-En}   & \textcolor{black}{En-Ja}  & Es-En              & Es-En         \\ \hline
    Training        & 209,941  & 180,853 & 145,503 & 228,697  & 48,301,352         & 238,872       \\
    Tuning          & 887      & 887     & 934     & 871     & 1,000              & 2,000         \\
    Test            & 5,473    & 4,706   & 1,664   & 8,469   & 1,000              & 14,522        \\ \hline
    \end{tabular}
    \label{tab:stat}
\end{table*}

% 预处理
\subsection{Data Preprocessing}
The English and Spanish datasets are tokenized by \emph{tokenizer.perl} and truecased by \emph{truecase.perl} provided by MOSES\footnote{\url{https://github.com/moses-smt/mosesdecoder}}, a statistical machine translation system proposed by \cite{Koehn2007moses}. 
The Chinese corpus is tokenized by \emph{Jieba} Chinese text segmentation\footnote{\url{https://github.com/fxsjy/jieba}}. 
Words in sentences are segmented into subwords by Byte-Pair Encoding (BPE) \cite{sennrich2016bpe} with 32k BPE operations.

% 模型配置
\subsection{Model Configuration}
\label{sec:config}
We use the Transformer proposed by Vaswani \etal \cite{vaswani2017attention} as our baseline and implement our work using the THUMT, an open-source toolkit for NMT developed by the Natural Language Processing Group at Tsinghua University \cite{zhang2017THUMT}\footnote{\url{https://github.com/thumt/THUMT}}.
We follow the configuration of the Transformer ``base model'' described in the original paper \cite{vaswani2017attention}.
Both encoder and decoder consist of 6 hidden layers each. 
All hidden states have 512 dimensions, eight heads for multi-head attention. The training batch contains about 6,520 source tokens, and we train the model about 200,000 training bathes.
We use the original regularization and optimizer in Transformer \cite{vaswani2017attention}. 
Finally, we evaluate the performance of the model by BLEU score \cite{papineni2002bleu} using \emph{multi-bleu.perl} on the \emph{tokenized} text.

\section{\textcolor{black}{Results and Analysis}}
\renewcommand\arraystretch{1.1}
\begin{table*}[ht]
    \centering
    \caption{\textcolor{black}{BLEU scores on the different datasets.
    % The marks ``\dag" after scores indicate that the proposed methods were significantly better than the baseline Transformer at the significance level $p$-value$<$0.05~\cite{collins-koehn-kucerova:2005:ACL}. 
    The scores in bold indicate the best ones on the same dataset. The last column indicates the time cost of different models on the News dataset.}}
    \label{tab:result}
    \begin{tabular}{lccccccc}
    \hline
    \multirow{2}{*}{\textbf{Model}} & \multicolumn{4}{c}{\textbf{TED Talks}}          & \textbf{Subtitles} & \multicolumn{2}{c}{\textbf{News}} \\
    & Zh-En          & Es-En          & \textcolor{black}{Fr-En} & \textcolor{black}{En-Ja} & Es-En              & Es-En            & \textcolor{black}{Cost. (hours)}  \\ 
    \hline
    RNNSearch* 
    & 16.09          & 36.55          & 30.79          & 10.41          & 39.90              & 22.95            & 23.60          \\
    Transformer
    & 17.76          & 38.53          & 30.92          & 11.73          & 39.96              & 23.71            & 28.59          \\
    \textcolor{black}{RNN with Memory Network} \cite{maruf2018document}   
    & -              & -              & 22.00          & -              & -                  & -                & -              \\
    Context-aware Transformer \cite{voita2018context} 
    & 18.24          & 38.74          & 31.20          & 11.87          & 40.19              & 23.76            & 42.09          \\
    Transformer with HAN \cite{miculicich2018han}
    & 17.79          & 37.24          & -              & -              & 36.23              & 22.76            & -              \\
    Our model
    & \textbf{18.69} & \textbf{39.20} & \textbf{31.97} & \textbf{12.01} & \textbf{40.74}     & \textbf{24.40}   & 42.32          \\ 
    \hline
    \end{tabular}
\end{table*}

\subsection{Translation Performance}
We choose the previous $m =3$ sentences as the contextual memory and using the Contextual RNN method to merge the embeddings.
Table \ref{tab:result} demonstrates the BLEU scores for different models on multiple corpora. 
The baseline is a re-implemented attention-based NMT system RNNSearch* \cite{hinton2012rnn} and Transformer \cite{vaswani2017attention} using THUMT kit.
We also employ the Context-aware model \cite{voita2018context} on these datasets, when we set the contextual memory size $m=1$ and without the inter-sentence attention.
\textcolor{black}{The results of RNN with Memory Network \cite{maruf2018document} and HAN model \cite{miculicich2018han} are reported by the authors.}

\textcolor{black}{The results in Table \ref{tab:result} demonstrate that our proposed model significantly outperforms all the comparing models, especially, our model is significantly better than the baseline Transformer at significance level $p$-value$<$0.05. 
Our proposed model outperforms the RNNSearch* baseline by 2.60 BLEU point on the TED Talks (Zh-En) dataset, 2.65 BLEU point on the TED Talks (Es-En) dataset, 1.18 BLEU point on the TED Talks (Fr-En) dataset, 1.60 BLEU point on the TED Talks (En-Ja) dataset, 0.84 BLEU point on the OpenSubtitles (Es-En) dataset, and 1.45 BLEU point on the WMT dataset (Es-En).}

\textcolor{black}{
Furthermore, our proposed model achieves the gains of 0.93, 0.67, 1.05, 0.28, 0.78, and 0.69 BLEU points on these four datasets individually over the Transformer baseline.}
Compared with the Context-aware Transformer proposed by \cite{voita2018context}, our proposed approach also raises the average 0.50 BLEU score on these different datasets. Moreover, the average increase of the BLEU score over the Transformer with HAN \cite{miculicich2018han} is 2.25 points.

\textcolor{black}{We note that the results on TED Fr-En are much higher than the result reported by Maruf and Haffai, and we deduce that it may be aroused by different prepossessing methods or BLEU styles.}

\textcolor{black}{The last column in Table \ref{tab:stat} indicates the time cost of different models on the News dataset (Es-En) under the same model setting mentioned in Section \ref{sec:config}. We can figure out that with the complexity of the model, the performance improves at the cost of running speed.}

\subsection{\textcolor{black}{Translation Analysis}}

% A-1: 补充针对 consistency, disambiguation, and coherence的指标分析

% A-3: more manual evaluation results on the three aspects, (1) ambiguous word auto-detection, (2) translation auto-checking

\textcolor{black}{To reflect the improvements of our proposed model more exactly, we will analyze the overall performance from three aspects mentioned above:  consistency, disambiguation, and coherence.
The translation of HAN model \cite{miculicich2018han} for comparison is downloaded from Miculicich's GitHub\footnote{\url{https://github.com/idiap/HAN_NMT/tree/master/test_out}}. }

\subsubsection{\textcolor{black}{Consistency}}

\textcolor{black}{In our work, the contextual memory is able to store the contextual sentences and help the model refine the translation. Thus, we follow the previous work \cite{kuang2018cache}, and calculate the average number of words in generated translations which are also in the contextual sentences fed into the contextual memory.
During our calculating process, punctuations, stop words, and \emph{UNK} are removed from the contextual sentences and translations. Table \ref{tab:consistency} shows the results of consistency on TED datasets with the memory size $m=3$. As shown in Table 3, HAN and our memory method can improve translation consistency compared to the baseline, confirming the claim that document translation can improve consistency between sentences. Our method is clearly closer to the reference than HAN and the baseline, demonstrating that our memory method is a more powerful approach for enhancing translation consistency.
}

\renewcommand\arraystretch{1.1}
\begin{table}[ht]
    \centering
    \caption{\textcolor{black}{Consistency test on TED Zh$\leftrightarrow$En test sets.}}
    \label{tab:consistency}
    \begin{tabular}{lcccc}
    \hline
    \multirow{2}{*}{\textbf{Model}} & \multicolumn{2}{c}{\textbf{TED Zh$\rightarrow$En}} & \multicolumn{2}{c}{\textbf{TED Zh$\rightarrow$En}} \\
    & \textit{pre-3}              & \textit{next-3}            & \textit{pre-3}              & \textit{next-3}            \\ \hline
    Reference         & 1.22               & 1.23              & 1.21               & 1.21              \\
    Transformer model & 1.04               & 1.05              & 1.02               & 1.04              \\
    HAN model         & 1.04               & 1.06              & 1.03               & 1.04              \\
    \bf Our model         & \textbf{1.12}      & \textbf{1.15}     & \textbf{1.11}      & \textbf{1.13}     \\ \hline
\end{tabular}
\end{table}

% TODO: ambiguous word auto-detection
\subsubsection{\textcolor{black}{Disambiguation}}
\textcolor{black}{We also want to investigate the ability of the word disambiguation of our model. 
We download the English-to-Chinese dictionary from free dictionary project\footnote{\url{https://www.dicts.info/}}, and select the words with multiple translation words in the source language to build a new dict $dict = \{{word^{src}: trans_1^{tgt},\ldots,trans_n^{tgt}\}}$. When the token $w$ in the source sentence and $ w \in \{word^{src}\}$, we count the appearance of the translation words $\{trans^{tgt}\}$ of $w$ in the corresponding translation sentence. }
\textcolor{black}{We argue that if a model is weak at disambiguation, to translate an ambiguous word with multiple word senses, the model would prefer one of the senses with the highest probability. The other corresponding candidate words' appearance will decrease accordingly. Thus, we use the Standard Deviation to evaluate the disambiguation ability.}

\renewcommand\arraystretch{1.1}
\begin{table}[ht]
    \centering
    \caption{\textcolor{black}{Disambiguation ability test on TED Zh$\rightarrow$En and Es$\rightarrow$En test sets.}}
    \label{tab:disam}
    \begin{tabular}{lcc}
    \hline
    \textbf{Model}    & \textbf{TED Zh$\rightarrow$En} & \textbf{TED Es$\rightarrow$En} \\ \hline
    Reference         & 652.15             & 541.16             \\
    Transformer model & 2691.81            & 2143.84            \\
    HAN model         & 2059.89            & 1457.06            \\
    \bf Our model (\textit{pre-3}) & \bf 1380.87            & \bf 1060.50            \\ \hline
    \end{tabular}
\end{table}

\textcolor{black}{Table \ref{tab:disam} illustrates the results of the disambiguation ability of different models. First, comparing the Transformer baseline and reference, it can be seen that the lower the standard deviation, the better the disambiguation will be. Second, compared with the baseline transformer, HAN decrease this metric on the two datasets. At the same time, our model achieved the lowest deviation value, indicating that the introduction of document information can alleviate the translation variety, i.e., have a disambiguation effect.}

% TODO: 分析句子相似度来表征Coherence, kuang2018cache
\subsubsection{\textcolor{black}{Coherence}}
\textcolor{black}{To further study how our proposed context-aware neural model improves the coherence in document translation, we follow the work of Lapata and Barzilay \cite{lapata2005automatic} to measure coherence as sentence similarity.
We represent each sentence as the mean of the distributed vectors of its words. Then, the similarity between the two sentences is determined by the cosine of their means.
For a fair comparison, we use the pre-trained language model BERT \cite{devlin-etal-2019-bert} to get the distributed vectors of words. 
}

\renewcommand\arraystretch{1.1}
\begin{table}[ht]
    \centering
    \caption{\textcolor{black}{Coherence test on TED Es$\rightarrow$En, Es$\rightarrow$En, and Subtitles Zh$\rightarrow$En test sets.}}
    \label{tab:coherence}
    \begin{tabular}{lccc}
    \hline
    \multirow{2}{*}{\textbf{Model}} & \multicolumn{2}{c}{\textbf{TED Talks}} & \textbf{Subtitles} \\
           & Zh$\rightarrow$En              & Es$\rightarrow$En             & Es$\rightarrow$En              \\ \hline
    Reference            & 0.67               & 0.67              & 0.60               \\
    Transformer model    & 0.62               & 0.61              & 0.54               \\
    HAN model            & 0.65               & 0.65              & 0.60               \\
    \bf Our model            & \textbf{0.66}      & \textbf{0.66}     & \textbf{0.60}      \\ \hline
    \end{tabular}
\end{table}

\textcolor{black}{Table \ref{tab:coherence} summarizes the comparison results. The Transformer baseline without document information has the lowest coherence score, while our system outperforms the HAN model slightly. On the one side, it demonstrates that both HAN and our model can improve the translation coherence, which leverages document features; on the other hand, it shows that our approach has certain advantages over HAN.}

% 例子 

\renewcommand\arraystretch{1.1}
\begin{table*}[ht]
    \centering
    \caption{Example of the translation result. The context sentences are three previous sentences before the source sentence and we use the Contextual RNN method to merge contextual argument embedding. The words in blue from context indicate the heuristic clues for better translation and the sentences in Chinese have been provided with English translation.}
    \begin{tabular}{l|p{10cm}}
    \hline
    Context sentence $\mathbf{c}_3$ & 它\ \textcolor{black}{去年}\ 秋天\ 遭遇\ 破产\ 因为\ 他们\ 遭到\ 入侵\ 。  \\
                           & \textit{(it was running into bankruptcy \textcolor{black}{last fall} because they were hacked into .)}                                                                                                                     \\
    Context sentence $\mathbf{c}_2$ & 有人\ 闯进去\ 彻底\ 毁\ 了\ 它\  \\
                           & \textit{(somebody broke in and they hacked it thoroughly .)} \\
    Context sentence $\mathbf{c}_1$ & 我\ \textcolor{black}{上周}\ 在\ 与\ 荷兰政府\ 代表\ 开会\ 时\ 问过\ ，\ 我\ 问\ 一位\ 领导\ 是否\ 他\ 发现\ 有\ 可能\ 有人\ 会\ 因为\ Diginotar\ 攻击\ 而\ 死亡\ 。\\
                           & \textit{(and I asked \textcolor{black}{last week} in a meeting with Dutch government representatives, I asked one of the leaders of the team whether he found plausible that people died because of the DigiNotar hack .)} \\ \hline
    Source sentence        & 他 的 回答 是 肯定 的 。                                                                                                                                                                                         \\ \hline
    Reference sentence     & \textit{and his answer was yes .}  \\
    Transformer model      & \textit{his answer is yes .}   \\
    HAN model              & \textit{his answer \textcolor{red}{and} yes .} \\
    \textbf{Our model}     & \textit{\textcolor{red}{and} his answer \textcolor{red}{was} yes .}                                                                                                                                                                       \\ \hline
    \end{tabular}
    \label{tab:example}
\end{table*}

\subsubsection{\textcolor{black}{Case Study}}
\paragraph{Example on Chinese-to-English}
We extract the 4,123-th parallel lines from TED Talks (Zh-En) and the contextual memory consists of three previous sentences before the source sentence. The the final contextual attention embedding $\mathbf{a}$ is merged by Contextual RNN method.
Table \ref{tab:example} shows an example from the TED Talks (Zh-En), on which the translation of our model is compared to other methods.
% The translation of HAN model is downloaded from \cite{miculicich2018han}'s GitHub\footnote{\url{https://github.com/idiap/HAN_NMT/tree/master/test_out}}. 
\textcolor{black}{This example shows that our proposed model can recognize the tense and even discourse relation from the document-level context and enhance the translation to more consistent and coherent. }

\paragraph{\textcolor{black}{Example on English-to-Japanese}}

\textcolor{black}{We select the 7,160-th parallel lines from TED Talks (En-Ja) test set and list the three previous sentences in the contextual memory. Compared with the baseline, for the word \emph{figure} with multiple word senses, our proposed model could recognize the correct word sense \emph{person} instead \emph{number}, and the attribute \emph{tragic} is also translated correctly. We infer that the word \emph{guy} in the context sentence $\mathbf{c}_3$ provides the translation clue, and it verifies that the contextual memory network enhances the disambiguation ability of our model.
}
\begin{CJK*}{UTF8}{min}
\renewcommand\arraystretch{1.1}
\begin{table*}[ht]
    \centering
    \caption{\textcolor{black}{We select the 7,160-th parallel lines from TED Talks (En-Ja) test set and list the three previous sentences in the contextual memory. Compared with the baseline, for the word \emph{figure} with multiple word senses, our proposed model could recognize the correct word sense \emph{person} instead \emph{number}, and the attribute \emph{tragic} is also translated correctly. We infer that the word \emph{guy} in the context sentence $\mathbf{c}_3$ provides the translation clue, and it verifies that the contextual memory network enhances the disambiguation ability of our model.}}
    \begin{tabular}{l|p{10cm}}
    \hline
    Context sentence $\mathbf{c}_3$ & this is a \textcolor{black}{guy} called e.p . \\
                           
    Context sentence $\mathbf{c}_2$ & the worst memory in the world .  \\
                           
    Context sentence $\mathbf{c}_1$ & his memory was so bad , that he didn \&apos;t even remember he had a memory problem , which is amazing .\\
    
    Source sentence        & and he was this incredibly \emph{tragic figure} , but he was a window into the extent to which our memories make us who we are . \\ \hline
    Reference sentence     & \textit{とても\ 悲劇\ 的\ な\ 人物\ です\ が\ どの\ 程度\ 記憶\ が\ 我々\ を\ 形作っ\ て\ いる\ か\ を\ 知る\ 手がかり\ と\ なる\ 存在\ です}\ \ \\
    Transformer\ model     & \textit{彼\ は\ 本当\ に\ \textcolor{red}{悲惨\ な\ 数字\ (disastrous number)}\ でし\ た\ が\ 私\ たち\ の\ 記憶\ が\ 私\ たち\ を\ どう\ 解釈\ する\ か\ に\ 窓\ を\ つけ\ て\ い\ まし\ た}   \\
    \textbf{Our model}     & \textit{彼\ は\ 非常\ に\ \textcolor{red}{悲劇\ 的\ な\ 人物\ (tragic person)}\ でし\ た\ が\ 私\ たち\ の\ 記憶\ が\ 私\ たち\ を\ どの\ よう\ に\ する\ か\ に\ つい\ て\ の\ 窓\ だっ\ た\ の\ です}                                                                                                                                                                       \\ \hline
    \end{tabular}
    \label{tab:example}
\end{table*}
\end{CJK*}

\renewcommand\arraystretch{1.1}
    \begin{table*}[ht]
    \centering
    \caption{The results on TED Talks with the different recurrent cores of Contextual RNN}
    \label{tab:core}
    \begin{tabular}{lcccccc}
    \hline
    \multirow{2}{*}{\textbf{Core}} & \multicolumn{3}{c}{\textbf{TED-Zh-En}} & \multicolumn{3}{c}{\textbf{TED-Es-En}} \\
     & forward  & backward  & bi-directional  & forward  & backward  & bi-directional  \\ \hline
    RNN           & 18.67    & 18.55     & 18.57           & 39.20    & 39.28     & 39.24           \\
    LSTM          & 18.58    & 18.67     & 18.56           & 39.21    & 39.27     & 39.21           \\
    GRU           & 18.54    & 18.63     & 18.61           & 39.19    & 39.23     & 39.20           \\ \hline
    \end{tabular}
\end{table*}

\section{Ablation Study}

% 比较RNN/LSTM/GRU
\subsection{\textcolor{black}{Effect of Recurrent Core of Contextual RNN}}
\textcolor{black}{In our proposed model, we use Contextual RNN to integrate the contextual information. Its recurrent core can also be replaced by GRU and LSTM with different styles. Table \ref{tab:core} illustrates the results when we change the recurrent core. We can observe that the recurrent core alteration influences our model slightly, and forward RNN is most efficient from the results.}

\subsection{Effect of Embedding Merging}
We choose the different embedding merging ways introduced in Section.\ref{sec:merging} to produce the final contextual attention embedding and compare the model performance on the different datasets with contextual memory size $m = 3$.

Table \ref{tab:merging} demonstrates the BLEU scores of the different embedding merging methods, and the model performs best on these datasets by contextual RNN merging method. 

\begin{table*}[ht]
    \centering
    \caption{\textcolor{black}{BLEU scores on the different datasets with various embedding merging ways.}}
    \begin{tabular}{lccccc}
        \hline
        \multirow{2}{*}{\textbf{Embedding Merging}} & \multicolumn{2}{c}{\textbf{TED Talks}} & \textbf{Subtitles} & \multicolumn{2}{c}{\textbf{News}}\\
        & Zh-En & Es-En & Es-En & Es-En & \textcolor{black}{Cost. (hours)}   \\ 
        \hline
        Concatenation    & 18.19          & 38.9          & 40.14          & 23.71         & 30.01          \\
        Average          & 18.23          & 38.95         & 40.34          & 23.82         & 37.37          \\
        Weighted Average & 18.44          & 39.16         & 40.68          & 24.37         & 39.58          \\
        Flat             & 18.48          & 39.15         & 40.59          & 24.33         & 32.59          \\
        Contextual RNN   & \textbf{18.67} & \textbf{39.2} & \textbf{40.74} & \textbf{24.4}  & 42.32          \\
        \hline
    \end{tabular}
    
    \label{tab:merging}
\end{table*}

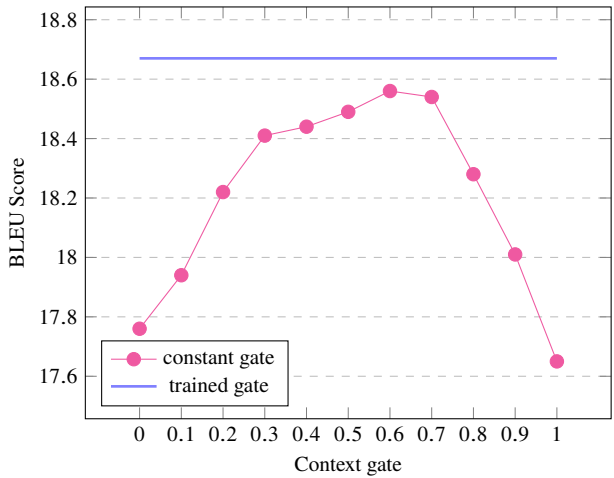
\begin{figure}[ht]
    \centering
    \begin{tikzpicture}
    \begin{axis}[
        xlabel={Context gate},
        ylabel={BLEU Score},
        xmin=0, xmax=1,
        ymin=17.6, ymax=18.7,
        xtick={0,0.1,0.2,0.3,0.4,0.5,0.6,0.7,0.8,0.9,1},
        ytick={17.6,17.8, 18.0, 18.2, 18.4, 18.6, 18.8},
        legend pos=south west,
        ymajorgrids=true,
        grid style=dashed,
        width=8.5cm,
        height=7cm,
        enlargelimits=0.13, 
        ]
        \addplot[
            color=magenta!80,
            mark=*,
            mark size=2.5pt,
            ]
            coordinates{
            (0,17.76)(0.1,17.94)(0.2,18.22)(0.3,18.41)(0.4,18.44)(0.5,18.49)(0.6,18.56)(0.7,18.54)(0.8,18.28)(0.9,18.01)(1,17.65)
            };
        \addlegendentry{constant gate}
        
        \addplot[
            color=blue!50,
            line width=1pt
            % mark=.,
            % mark size=2.5pt,
            ]
            coordinates{
            (0,18.67)(1,18.67)
            };
        \addlegendentry{trained gate}
          
  \end{axis}
  \end{tikzpicture}
  \caption{Results on TED Talks (Zh-En) dataset with different context gating ways.}
  \label{fig:gating}
\end{figure}

\subsection{Effect of Context Gating}
We also investigate the impact of context gate $g$ by using the different given constants and compare the results with the context gate trained by the model. For instance, if the context gate $g$ equals 0, the model is the vanilla Transformer model, and context gate $g=1$ means the model only encodes the final contextual attention embedding $\mathbf{a}$ from the context sentence(s) without the source attention.
Fig. \ref{fig:gating} illustrates the performance of the different context gate values when the contextual memory size $m = 3$.
Of course, the context gate obtained from the model performs better than the fixed context gate, and meanwhile, both source information and context information are essential to the model.

\subsection{Effect of Contextual Information}

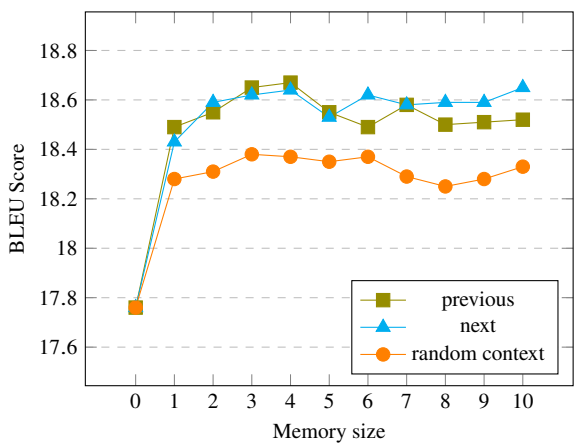
\begin{figure}[ht]
    % \centering
    % \begin{minipage}{0.45\textwidth}
    \centering
    \begin{tikzpicture}
    \begin{axis}[
        xlabel={Memory size},
        ylabel={BLEU Score},
        xmin=0, xmax=10,
        ymin=17.6, ymax=18.8,
        xtick={0,1,2,3,4,5,6,7,8,9,10},
        ytick={17.6, 17.8, 18.0, 18.2, 18.4, 18.6, 18.8},
        legend pos=south east,
        ymajorgrids=true,
        grid style=dashed,
        width=8cm,
        height=6.5cm,
        enlargelimits=0.13, 
        ]
        \addplot[color=olive,mark=square*,mark size=2.5pt,]
            coordinates{(0,17.76)(1,18.49)(2,18.55)(3,18.65)(4,18.67)(5,18.55)(6,18.49)(7,18.58)(8,18.50)(9,18.51)(10,18.52)};
        \addlegendentry{previous}
        
        \addplot[color=cyan,mark=triangle*,mark size=3pt,]
            coordinates{(0,17.76)(1,18.43)(2,18.59)(3,18.62)(4,18.64)(5,18.53)(6,18.62)(7,18.58)(8,18.59)(9,18.59)(10,18.65)};
        \addlegendentry{next}
        
        \addplot[color=orange,mark=*,mark size=2.5pt,]
            coordinates{(0,17.76)(1,18.28)(2,18.31)(3,18.38)(4,18.37)(5,18.35)(6,18.37)(7,18.29)(8,18.25)(9,18.28)(10,18.33)};
        \addlegendentry{random context}
    \end{axis}
    \end{tikzpicture}
    \caption{Results on TED Talks (Zh-En) dataset with different contextual memory size $m$ and different context selection.}
    \label{fig:memory_size}
    % \end{minipage}
\end{figure}

\paragraph{Different context sentence definition}
The context sentences in our work are the previous three sentences of the current sentence. We investigate the effect of the different context sentence definition on the TED Talks (Zh-En) dataset. Following the work of Context-aware Transformer \cite{voita2018context}, we use the previous sentence(s), next sentence(s) and the random selected context sentence(s) form the document as the context sentence(s).
As shown in Fig.\ref{fig:memory_size},  the model which uses the previous sentence(s) as the context sentence(s) could get the best performance on the TED Talks (Zh-En) dataset, and it is in agreement with the work of the Context-aware Transformer \cite{voita2018context}.

\paragraph{Different contextual memory size}
We also compare the effect with the different contextual memory size $m$ on the TED Talks (Zh-En) dataset. If the contextual memory size $m$ equals 0, the model is the original Transformer model.
As shown in Fig.\ref{fig:memory_size}, more contextual information appears beneficial to model translation, and the BLEU score gets better with more context sentences. However, it changes slightly when the contextual memory size $m$ greater than 4.

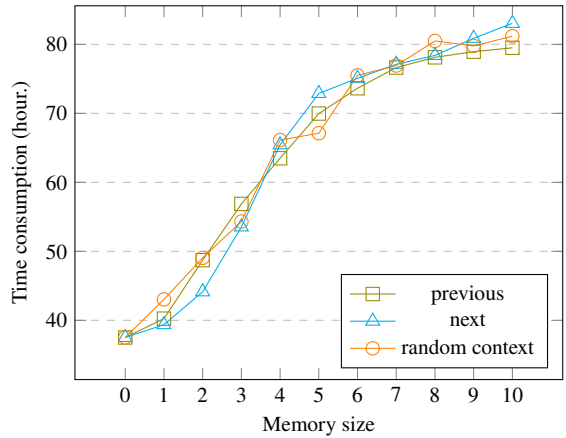
\begin{figure}[ht]
    % \begin{minipage}{0.45\textwidth}
    \centering
    \begin{tikzpicture}
    \begin{axis}[
        xlabel={Memory size},
        ylabel={Time consumption (hour.)},
        xmin=0, xmax=10,
        ymin=37, ymax=80,
        xtick={0,1,2,3,4,5,6,7,8,9,10},
        ytick={30, 40, 50, 60, 70, 80},
        legend pos=south east,
        ymajorgrids=true,
        grid style=dashed,
        width=8cm,
        height=6.5cm,
        enlargelimits=0.13, 
        ]
        \addplot[color=olive,mark=square,mark size=2.5pt,]
            coordinates{(0,37.51)(1,40.24)(2,48.72)(3,56.87)(4,63.49)(5,69.97)(6,73.64)(7,76.63)(8,78.12)(9,78.93)(10,79.50)};
        \addlegendentry{previous}
        
        \addplot[color=cyan,mark=triangle,mark size=3pt,]
            coordinates{(0,37.51)(1,39.37)(2,44.13)(3,53.54)(4,65.44)(5,72.88)(6,75.08)(7,77.11)(8,78.39)(9,80.88)(10,83.04)};
        \addlegendentry{next}
        
        \addplot[color=orange,mark=o,mark size=2.5pt,]
            coordinates{(0,37.51)(1,43.03)(2,49.05)(3,54.32)(4,66.13)(5,67.12)(6,75.50)(7,76.93)(8,80.49)(9,79.77)(10,81.18)};
        \addlegendentry{random context}
  \end{axis}
  \end{tikzpicture}
  \caption{The comparison of time consumption with different contextual memory size $m$ and different context selection.}
  \label{fig:time}
% \end{minipage}
\end{figure}

\subsection{Time Consumption}
We also compare time consumption with different contextual memory sizes and context definitions mentioned in the above section, and we illustrate Figure \ref{fig:time} according to the statistics.
The original model needs 17.76 hours, and obviously, the proposed method needs more hours for 200 thousand steps on the TITAN RTX GPU device. Our proposed model needs a bit more time because the contextual associated memory network is more complex and has to train more hyper-parameters during the training process.

\section{Related Work}
% 已有方法，可以分为两类: 1)如何引入文档信息 2)引进怎样的文档信息;
The existing work about NMT on the document-level can be divided into two parts: one is how to obtain the document-level information in NMT, and the other is how to integrate the document-level information.
% The existing work about NMT on the document-level can be divided by context integrating methods.

\subsection{Mining Document-level Information}
Tiedemann \etal \cite{tiedemann2017context} merely concatenate the context in two ways: (1) extending the source sentence, which includes the context from the previous sentences in the source language, and (2) extending translation units, which increase the segments to translate. 

Michel \etal \cite{michel2018extreme} propose a simple yet parameter-efficient adaption method that only requires adapting the \emph{Specific Vocabulary Bias} of output softmax to each particular use of the NMT system and allows the model to reflect distinct linguistic variations through translation better.

Mac \etal \cite{Mac2019UsingWD} present a \emph{Word Embedding Average} method to add source context that captures the whole document with accurate boundaries, taking every word into account by an averaging method. 

Kang \etal \cite{kang2020dynamic} propose to select dynamic context so that the document-level translation model can utilize the more useful selected context sentences to produce better translations via reinforcement learning.

\subsection{Integrating Document-level Information}

\paragraph{Gating Context}
The context gate can automatically control the ratios of source and context representations contributions to the generation of target words \cite{tu2017context}.
Wang \etal \cite{wang2017exploiting} introduce this mechanism in their work to dynamically control the information flowing from the global text at each decoding step.
Kuang \etal \cite{kuang2018gate} propose an inter-sentence gate model, which is based on the attention-based NMT and uses the same encoder to encode two adjacent sentences and controls the amount of information flowing from the preceding sentence to the translation of the current sentence with an inter-sentence gate.
% This gate framework assigns element-wise weights to the input signals, calculated by the context vectors of two adjacent sentences, target word representation, and the decoder hidden state.

\paragraph{Document RNN}
Wang \etal \cite{wang2017exploiting} propose a cross-sentence context-aware RNN approach to produce a global context representation called Document RNN.
Given a source sentence in the document to be translated and its $m$ previous sentences, they can obtain all sentence-level representations after processing each sentence.
The last hidden state represents the summary of the whole sentence as it stores order-sensitive information.
The last hidden state represents the summary of the global context over the sequence of the above sentence-level representations.

\paragraph{Cache-based Neural Model}
Tu \etal \cite{tu2018learning} propose to augment the NMT models with an external cache to exploit translation history. At each decoding step, the probability distribution over generated words is updated online depending on the translation history retrieved from the cache with a query of the current attention vector, which helps NMT models adapt over time dynamically.

\paragraph{Context-Aware Transformer Model}
Voita et al. \cite{voita2018context} introduce the context information into the Transformer \cite{vaswani2017attention} and leave the Transformer's decoder intact while processing the context information on the encoder side. The model calculates the gate from the source sentence attention and the context sentence attention, exploiting their gated sum as the encoder output.

Zhang \etal \cite{zhang2018improving} also extend the Transformer with a new context encoder to represent document-level context while incorporating it into both the original encoder and decoder by multi-head attention. 

Miculicich \etal \cite{miculicich2018han} propose a  \emph{Hierarchical Attention Networks (HAN) NMT} model to capture the context in a structured and dynamic pattern. Each predicted word uses word-level and sentence-level abstractions and selectively focuses on different words and sentences.

Tan \etal \cite{tan-etal-2019-hierarchical} propose a \emph{hierarchical modeling of global document context model} to improve document-level translation, which is hierarchically extracted from the entire global text with a sentence encoder to model intra-sentence information and a document encoder to model document-level inter-sentence context representation.

Ma \etal \cite{ma-etal-2020-simple} propose a \emph{Flat-Transformer model} with a simple and effective unified encoder that model the bi-directional relationship between the contexts and the source sentences.

Chen \etal \cite{chen-etal-2020-modeling} propose to improve document-level NMT by the means of discourse structure information, and the encoder is based on a HAN \cite{miculicich2018han}. They parse the document to obtain its discourse structure, then introduce a Transformer-based path encoder to embed the discourse structure information of each word and combine the discourse structure information with the word embedding. 

Most of the previous works only focus on integrating context embedding or considering the context selection, but our work can mine the most related part among the contextual memory at each step.

% 结论 

\section{Conclusion and Future Work}
We propose a memory network enhancement over Transformer-based NMT, which provides a natural solution for modeling the detailed document-level context. 
Experiments show that our model performs better on the datasets of multiple domains and language pairs and can capture salient document-level contextual clues, select the most relevant part related to the input sequence from the contextual memory, and effectively enhance strong NMT baselines. 
% Comparing to the latest state-of-the-art document-level NMT solution, our proposed memory network modeling brings even higher performance-boosting over strong baselines, which firmly verifies the effectiveness of the proposed method.

% 在未来的工作中，我们考虑基于当前框架进一步引入更丰富的discourse信息增强NMT, 由于discource信息的注入会带来大量噪音的，另一方面，discourse信息有可能特别大，长，内部结构复杂，因此，需要有效抽象其关键特征信息，discource信息部分提供了启发式特征，会改进训练和decoding效果。
We will consider better context selection in our future work, like using discourse information to enhance our model. On the one hand, the discourse information will provide the heuristic, but on the other hand, it will bring much noise, and the internal structure may be incredibly complicated. Therefore, it is necessary to abstract its critical feature information effectively.

\bibliographystyle{ieicetr}% bib style
\bibliography{mem}% your bib database

% \begin{thebibliography}{99}% more than 9 --> 99 / less than 10 --> 9
% \bibitem{}
% \end{thebibliography}

%\profile*{}{}% without picture of author's face

\end{CJK*}
\end{document}